\documentclass[11pt]{article}
\usepackage{eamt20}
\usepackage{times}
\usepackage{url}
\usepackage{latexsym}
\usepackage[small,bf]{caption} 
\usepackage{todonotes} 
\usepackage{soul} 

\usepackage{array} 
\usepackage{CJKutf8}
\usepackage{multirow}
\usepackage{amssymb}

\usepackage{graphicx}
\graphicspath{ {./images/} }

\title{Fine-grained Human Evaluation of Transformer and Recurrent Approaches to Neural Machine Translation for English-to-Chinese}

\author{Yuying Ye\\
Digital Humanities Programme\\
University of Groningen\\
  The Netherlands\\  {\tt y.ye.yuying@gmail.com} \And Antonio Toral\\
Center for Language and Cognition\\
University of Groningen\\
  The Netherlands\\  {\tt a.toral.ruiz@rug.nl}}

\date{}

\begin{document}
\begin{CJK*}{UTF8}{gbsn}
\maketitle
\begin{abstract}

This research presents a fine-grained human evaluation to compare the Transformer and recurrent approaches to neural machine translation (MT), on the translation direction English-to-Chinese.
To this end, we develop an error taxonomy compliant with the Multidimensional Quality Metrics (MQM) framework
that is customised to the relevant phenomena of this translation direction.
We then conduct an error annotation using this customised error taxonomy on the output of state-of-the-art recurrent- and Transformer-based MT systems on a subset of WMT2019's news test set.
The resulting annotation shows that, compared to the best recurrent system, the best Transformer system results in a 31\% reduction of the total number of errors and it produced significantly less errors in 10 out of 22 error categories. 
We also note that two of the systems evaluated do not produce any error for a category that was relevant for this translation direction prior to the advent of NMT systems: Chinese classifiers.


\end{abstract}

\section{Introduction}

The field of machine translation (MT) has been revolutionised in the past few years by the emergence of a new approach: neural MT (NMT). 
NMT is a dynamic research area and we have witnessed two mainstream architectures already, the first of which is based on recurrent neural networks (RNN) with attention \cite{bahdanau_neural_2014} while the second, referred to as Transformer, makes use of the self-attention mechanism in non-recurrent networks \cite{vaswani-attention-2017}.

Several studies have analysed in depth, using both automatic and human evaluation methods, the resulting translations of NMT systems under the recurrent architecture and compared them to the translations of the previous mainstream approach to MT: statistical MT \cite{koehn-smt-2003}, e.g. \cite{bentivogli_neural_2016,castilho-2017,klubicka_quantitative_2018,popovic2017comparing,shterionov2018human}. 
However, while the Transformer architecture has brought, at least when trained with sufficient data, considerable gains over the recurrent architecture \cite{vaswani-attention-2017},
the research conducted to date that analyses the resulting translations of these two neural approaches is, to the best of our knowledge, limited to automatic approaches~\cite{burlot-etal-2018-wmt18,lakew-etal-2018-comparison,tang2018self,tang-etal-2018-analysis,tran-etal-2018-importance,yang-etal-2019-assessing}.



In this paper we conduct a detailed human analysis of the outputs produced by state-of-the-art recurrent and Transformer NMT systems. Namely, we manually annotate the errors found according to a detailed error taxonomy which is compliant with the hierarchical listing of issue types defined as part of the Multidimensional Quality Metrics (MQM) framework~\cite{lommel-mqm-2014}.
We carry out this analysis for the news domain in the English-to-Chinese translation direction.
To this end, we define an error taxonomy that is relevant to the problematic linguistic phenomena of this translation direction.
This taxonomy is then used to annotate errors produced by NMT systems that fall under the recurrent and Transformer architectures. 

The main contributions of this paper can then be summarised as follows:
\begin{enumerate}
    \item We develop an MQM-compliant error taxonomy tailored to the English-to-Chinese translation direction.
    \item We conduct, to the best of our knowledge, the first 
    human fine-grained error analysis of Transformer-based versus recurrent NMT.
\end{enumerate}

The rest of the paper 
is arranged in the following way.
Section \ref{s:related_work} presents a brief review of related work. 
Next, Section \ref{s:mt} outlines the recurrent- and Transformer-based NMT systems and the dataset used in our experiments. 
Subsequently, Section \ref{s:error_annotation} 
presents the methodology for error annotation and the definition of the error taxonomy, followed by results and statistical analysis of the annotation. 
Finally, Section~\ref{s:conclusion} gives a conclusion and suggestions for future work.


\section{Related Work}\label{s:related_work}

This section provides an overview of related research on the two topics that correspond to our main contributions: human error analysis of MT outputs for the language pair English--Chinese (Section~\ref{s:related_analysis_enzh}) and analyses of MT systems based on the recurrent and Transformer architectures (Section~\ref{s:related_recurrent_transformer}).

\subsection{Human Error Analyses of MT for Chinese}\label{s:related_analysis_enzh}


One of the first taxonomies of MT errors, 
by Vilar et al.~\shortcite{vilar_error_2006}, had a specific error typology for the Chinese-to-English translation direction, in accordance with the specific relevant phenomena of this language pair.
Compared to their base taxonomy, a refined categorisation of word order was added to mark syntactic mistakes that appear in translations of questions, infinitives, declarative and subordinate sentences. 
In addition, 
the error type \emph{Unknown words} was refined into four sub-types: \emph{Person}, \emph{Location}, \emph{Organisation} and \emph{Other proper names}.


Li et al.~\shortcite{li_chinese_2009} carried out an error analysis for the Chinese-to-Korean translation direction with only three categories from the taxonomy of Vilar et al.~\shortcite{vilar_error_2006} (\emph{Missing words}, \emph{Wrong word order} and \emph{Incorrect words}), and  they replaced \emph{Incorrect words} with two more specific categories: one for both wrong lexical choices and extra words and another for wrong modality. The simplified taxonomy was used to check if their method of reordering verb phrases, prepositional phrases and modality-bearing words in the Chinese data resulted in an improved MT system.


Hsu~\shortcite{hsu_error_2014} adapted the classification scheme of 
Farrús et al.~\shortcite{farrus-linguistic-2010} to conduct an error analysis for the Chinese-to-English translation direction.
The error taxonomy of Farrús et al.~\shortcite{farrus-linguistic-2010} was originally defined for Catalan$\rightarrow$Spanish. Its first level corresponded to five types of errors, related to different linguistic levels: orthographic, morphological, lexical, semantic and syntactic.



Castilho et al.~\shortcite{castilho-2017} assessed the output of two MT systems (statistical and recurrent) on patents, also for the Chinese-to-English translation direction.
For this, they used a custom error taxonomy consisting of the error types \emph{Punctuation}, \emph{Part of speech}, \emph{Omission}, \emph{Addition}, \emph{Wrong terminology}, \emph{Literal translation}, and \emph{Word form}.

Hassan et al.~\shortcite{hassan-parity-2018} analysed the output of a Transformer-based MT system, again for the Chinese-to-English translation direction, using a two-level taxonomy based on that by Vilar et al.~\shortcite{vilar_error_2006}.
The first level contains nine error types: \emph{Missing words}, \emph{Word repetition}, \emph{Named entity}, \emph{Word order}, \emph{Incorrect words}, \emph{Unknown words},
\emph{Collocation}, \emph{Factoid}, and \emph{Ungrammatical}.
Only the error type \emph{Named entity} has a second level, with five subcategories: \emph{Person}, \emph{Location}, \emph{Organisation}, \emph{Event}, and \emph{Other}.

As we can observe in these related works, fine-grained human evaluation for the English--Chinese language pair has been hitherto conducted, to the best of our knowledge, (i) only for the Chinese-to-English direction and (ii) with error taxonomies that were either developed prior to the advent of the MQM framework or that were designed ad-hoc and were not thoroughly motivated.
The position of our paper in these regards is thus clearly novel: (i) our analysis is for the English-to-Chinese translation direction and (ii) we devise and use an error taxonomy that is compliant with the MQM framework.

\subsection{Analyses of Recurrent versus Transformer MT Systems}\label{s:related_recurrent_transformer}

Tang et al.~\shortcite{tang2018self} compared recurrent- and Transformer-based MT systems on a syntactic task that involves long-range dependencies (subject-verb agreement) and on a semantic task (word sense disambiguation)
The recurrent system outperformed Transformer on the syntactic task while Transformer was better than the recurrent system on the semantic task.
The latter finding was corroborated by Tang et al.~\shortcite{tang-etal-2018-analysis}.



Tran et al.~\shortcite{tran-etal-2018-importance} compared the recurrent and Transformer architectures with respect to their ability to model hierarchical structure in a monolingual setting, by means of two tasks: subject-verb agreement and logical inference.
On both tasks, the recurrent system outperformed Transformer, slightly but consistently.


Burlot et al.~\shortcite{burlot-etal-2018-wmt18} confronted English$\rightarrow$Czech Transformer- and recurrent-based MT systems submitted to WMT2018\footnote{\url{http://www.statmt.org/wmt18/}} on a test suite 
that addresses 
morphological competence, based on the error typology by Burlot and Yvon~\shortcite{burlot-yvon-2017-evaluating}.
The recurrent system outperformed Transformer on cases that involve number, gender and tense, while both architectures performed similarly on agreement. It is worth noting that agreement here regards local agreement (e.g. an adjective immediately followed by a noun), while the aforementioned cases of agreement in which a recurrent system outperforms Transformer~\cite{tang2018self,tran-etal-2018-importance} regard long-distance agreement.


Yang et al.~\shortcite{yang-etal-2019-assessing}
assessed the ability of both architectures to learn word order.
When trained on a specific task related to word order, word reordering detection, a recurrent system outperformed Transformer.
However, when trained on a downstream task, MT,
Transformer was able to learn better positional information.

Lakew et al.~\shortcite{lakew-etal-2018-comparison}
evaluated multilingual NMT systems under the Transformer and recurrent architectures in terms of their morphological, lexical, and word order errors.
In both architectures lexical errors were found to be the most prominent ones, followed by morphological, and lastly come reordering errors.
The authors compared the number of errors in bilingual, multilingual and zero-shot systems, both for recurrent and Transformer, and found multilingual and zero-shot systems to be more competitive with respect to bilingual models for Transformer than for recurrent.








\section{Machine Translation Systems}\label{s:mt}


This section reports on the MT systems and the dataset used in our experiments.

We have used output from systems that fall under the recurrent and Transformer architectures and were top-ranked at the news translation shared task at the Conference on Machine Translation (WMT). 
We chose the University of Edinburgh's MT system~\cite{sennrich_edinburgh_2017} as our recurrent NMT system due to the fact that this system had the highest BLEU score (36.3) for the translation direction English$\rightarrow$Chinese at WMT2017\footnote{\url{http://www.statmt.org/wmt17/}} and it was ranked first (tied with other two systems) in the human evaluation.




As for the Transformer-based MT system used in our research, we have taken the PATECH submission to WMT2019.\footnote{\url{ http://matrix.statmt.org/systems/show/4243}}
We conducted our experiments before the human evaluation of WMT2019 was available, and therefore we chose the PATECH's system based on the automatic evaluation of WMT2019, in which this system was the best performing one.\footnote{\url{http://matrix.statmt.org/matrix/systems_list/1908}}
However, PATECH's system was not included in the human evaluation of WMT2019.
Therefore we carried out an additional annotation on the top-performing system from that human evaluation: the Transformer system developed by Kingsoft AI Lab~\cite{guo-etal-2019-kingsofts}, hereafter referred to as KSAI.



Before our human error analysis, we would like to compare the recurrent and Transformer MT systems in terms of an automatic evaluation metric.
This is not possible from their outputs since they correspond to two different test sets (\texttt{newstest2017} and \texttt{newstest2019}).
In order to be able to compare them, we asked the developer of the recurrent system to provide us with the output from their system for \texttt{newstest2019}.
As shown in Table \ref{tab:bleu}, the use of the Transformer architecture leads to a considerable improvement compared to the recurrent system (on average 31.4\% relative in terms of BLEU). 
While the gap between the two architectures 
is large based on BLEU, this is an overall metric and therefore does not provide any insight into which aspects of the translation have improved with Transformer with respect to the recurrent system.
To gain further insight we conduct a fine-grained human error analysis in the following section.

\begin{table}[ht!]
\centering
\small
\begin{tabular}{c| m{5em} | m{5em}}
\hline
 RNN & Transformer (PATECH) & Transformer (KSAI)
\\ \hline
33.1 & 44.6 & 42.4
\\ \hline
\end{tabular}
\caption{Automatic evaluation (BLEU scores) of the 3 MT systems on the WMT 2019 news test set.}
\label{tab:bleu}
\end{table}


\section{Error Annotation}\label{s:error_annotation}

This section details the annotation setup (Section \ref{s:annotation}),
explains how we defined our MQM-compliant 
error taxonomy adapted to the relevant characteristics of translating from English into Chinese and the challenges faced by NMT systems in this translation direction (Section \ref{s:taxonomy}) and presents the results of the annotation, as well as analysis and discussion thereof (Section \ref{s:result}). 


\subsection{Annotation Setup}\label{s:annotation}

We use \texttt{translate5},\footnote{\url{http://www.translate5.net}} an open-source web-based tool, as the annotation environment. \texttt{translate5} was installed 
on a cloud server, so that it could be accessed remotely by annotators.
The source text and reference translation are provided next to the NMT translations.

The annotation was performed by two annotators who are native Chinese speakers with fluent English. They both had an academic background and experience in translation. Prior to annotation, they were fully informed on the annotation environment and were provided with annotation instructions, comprising MQM's usage guidelines and decision tree~\cite{burchardt-guidelines-2014}.

The dataset used in our experiments is the test set from WMT2019 (\texttt{newstest2019}) for English$\rightarrow$Chinese. This test set is chosen due to the fact that we have outputs for the RNN- and Transformer-based MT systems (see Section~\ref{s:mt}), and also because it is a commonly-used benchmark in the MT community.
In our error annotation we use two subsets of this test set.
\begin{itemize}
    \item A calibration set, made of the first 40 sentences from the testset. 
    This refers to a small sample of annotation data that annotators work on
before the actual annotation task takes place. Its purpose 
is twofold: (i) we use it to find out which error types occur in the translations and therefore use it to guide the refinement of the error taxonomy in a data-driven way;
(ii) we also use it to identify disagreements between the annotators.
    \item An evaluation set, made up of 100 sentences from the test set. In order to have intersentential context, these sentences are taken from six documents (five full documents and the first sentences of the sixth document up to 100 sentences are reached).
    Using this evaluation set led then to the annotation of 
    500 sentences (100 distinct sentences times two MT systems (RNN and PATECH) times two annotators, plus the annotation of the 100 sentences for a third system (KSAI) by one annotator).
\end{itemize}


The annotators annotated the calibration set with our custom error taxonomy (see Figure~\ref{fig:chinese tagest}), after which they discussed difficult cases and reached agreement on how to annotate them. 
Then they 
annotated the translations of the evaluation set.
Once annotators started working on the evaluation set, they were not allowed to discuss problems in annotation any more. 

\subsection{Error Taxonomy}\label{s:taxonomy}
\begin{figure*}[htb]
\centering
\includegraphics[scale=0.6]{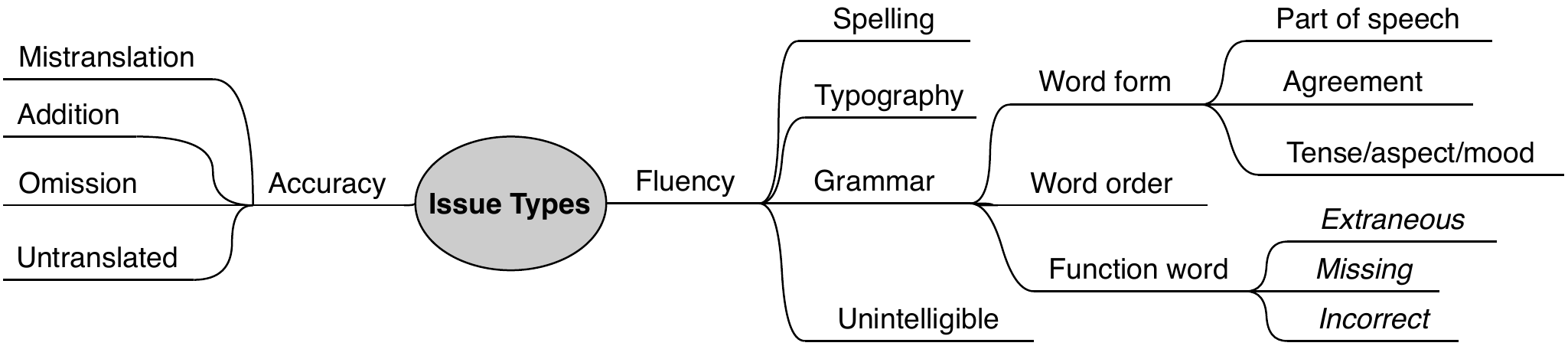}
\caption{The sample MQM-compliant error hierarchy for diagnostic MT evaluation. The italicised issue types are not included in the standard MQM issue types.}
\label{fig: standard-MQM-core}
\end{figure*}

We decided to develop our error taxonomy based on the MQM framework 
developed at the QTLaunchPad project~\cite{lommel-mqm-2014}, after reviewing different translation quality evaluation frameworks.
MQM stands out with its extensive standardised issue types\footnote{\url{http://www.qt21.eu/mqm-definition/issues-list-2015-12-30.html}} which are provided with clear definitions and explanations. 
In addition, a thorough guideline and decision tree\footnote{\url{http://www.qt21.eu/downloads/MQM-usage-guidelines.pdf}} are available to assist annotators.
Furthermore, this framework allows the building of customised error taxonomies.

Following the method of Klubička et al.~\shortcite{klubicka_quantitative_2018}, our customisation process started with the sample MQM-compliant hierarchy for diagnostic MT evaluation (Figure \ref{fig: standard-MQM-core}) as the initial stage of our error taxonomy. The sample MQM tagset went through the preliminary selection of issue types to be used for fine-grained MT evaluation.

\begin{figure*}[htb]
\centering
\includegraphics[scale=0.55]{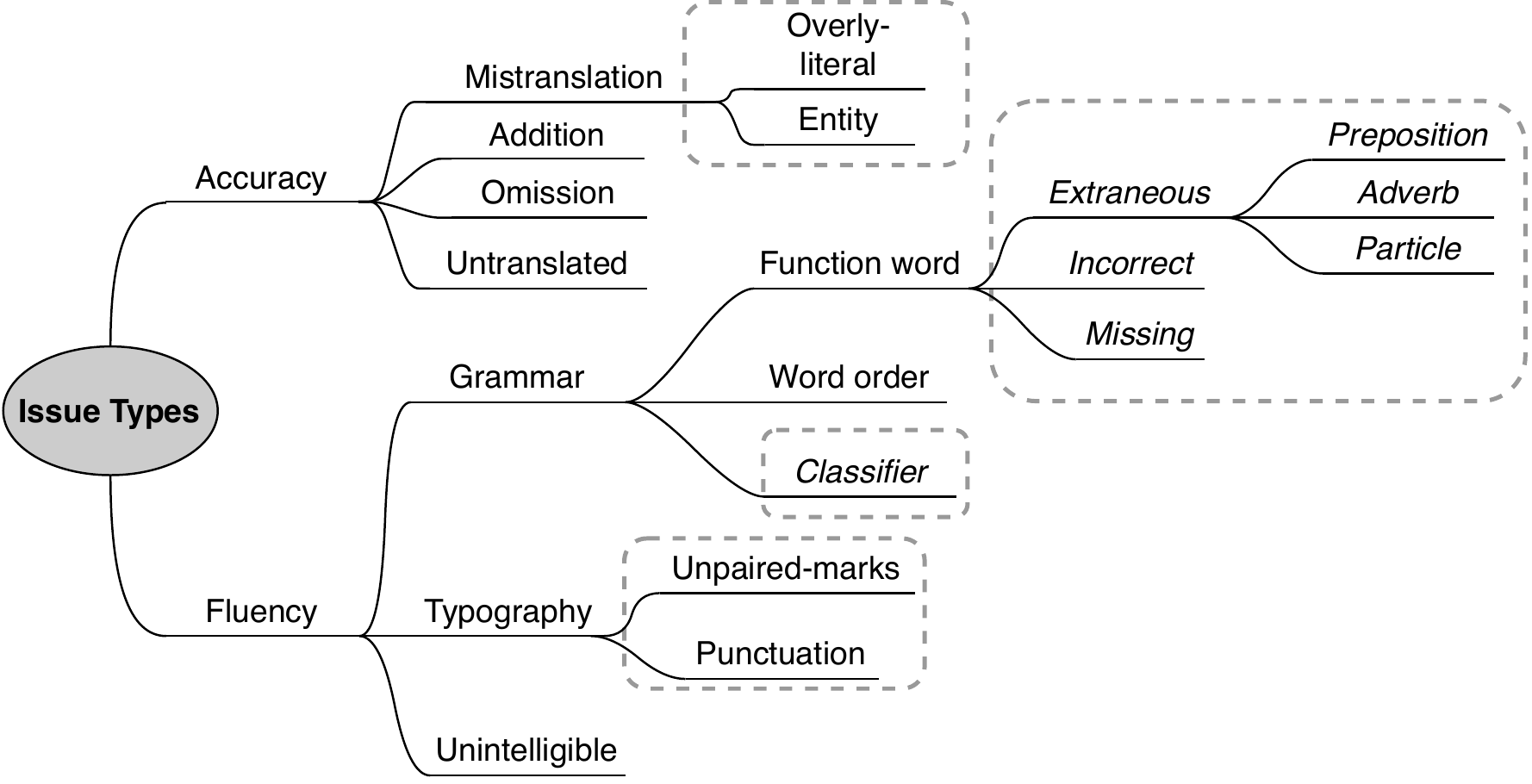}
\caption{The MQM-compliant error taxonomy for the translation direction English$\rightarrow$Chinese.
All the changes are marked by boxes with grey dotted lines and the issue types that are not included in the MQM issue types are italicised.}
\label{fig:chinese tagest}
\end{figure*}

We annotated the calibration set with the sample MQM-compliant hierarchy to find out what types of errors occur in the outputs of our MT systems. 
Based on the results of the calibration set, we defined the complete tagset (shown in Figure \ref{fig:chinese tagest}).
In the following subsections we provide detailed information concerning each of the modifications made to the error taxonomy.

\subsubsection{Word Form \& Spelling}
Given that Chinese is an analytic language without inflection and its writing system is logographic, the issue types \emph{Word form} and \emph{Spelling} are of no interest to our research agenda. 

\subsubsection{Classifier}
We add one of the distinctive features of Chinese part-of-speech, the usage of classifiers, which have been researched thoroughly in Chinese linguistics \cite{jin_partition_2018} and Chinese language processing \cite{huang_mandarin_2017}. In short, classifiers are special linguistic units located behind a number, demonstrative or certain quantifiers. These classifiers 
do not have a counterpart in English, 
which might give rise to translation problems. Examples of classifiers are shown in Table \ref{table:classifier}. How MT systems handle such a specific linguistic phenomenon is of interest to us. 

\begin{table} [ht!]
\centering
    \begin{tabular}{ c|c|c }
        \hline
        Pronoun & Classifier & Noun \\ \hline
        每(\emph{mei})&\underline{个}(\emph{ge})&角落(\emph{jiaoluo})\\
         Every &&corner\\
         \hline
         一(\emph{yi})&\underline{架}(\emph{jia})&飞机(\emph{feiji})\\
         One&&plane\\
         \hline
    \end{tabular}
    \caption{Examples of classifiers in Chinese. The classifiers are underscored.}
    \label{table:classifier}
\end{table}

\subsubsection{Typography}
We extend the issue type \emph{Typography} into two specific subtypes, based on the result of the calibration set. Though an unpaired quote 
or a misuse of punctuation is less likely to damage the comprehension of the content critically than other errors, as stated in Vilar et al.~\shortcite{vilar_error_2006}, the Chinese$\rightarrow$English error annotation conducted by Hsu~\shortcite{hsu_error_2014} shows that punctuation accounts for 10\% of the errors. Such a high amount of punctuation mistakes could be a nuisance in the MT output. Incorrect usage of \emph{Typography} could negatively influence the reception of a translation, since the reader might consider such an error as a sign of lack of professionalism, and therefore react by distrusting the content.

\subsubsection{Mistranslation}
Preliminarily, we observe that \emph{Mistranslation} is a major issue in the calibration set and that related translation errors \emph{Overly-literal} and \emph{Entity} appear frequently. We have thus decided to specify them as sub-types of \emph{Mistranslation}. 
Vilar et al.~\shortcite{vilar_error_2006} also included entity errors in their error typology for the Chinese$\rightarrow$English language pair and further divided them into specific sub-types. As their result showed, this issue type only amounted to a small percentage of the errors. Therefore, the issue type \emph{Entity} is not further specified in our taxonomy. 

\subsubsection{Function word}

\emph{Function word} is extended to one extra layer under \emph{Extraneous} with the intention of covering westernised Chinese expressions that were observed in the calibration set. Westernised Chinese refers to a cross-lingual phenomenon of imposing English grammar on Chinese, which is manifested in many problematic forms, abuse of function words especially \cite{xie-chinese-grammar-2001}. 
The relations between sentence parts, tenses and aspects are often shown through word order, particles or context in Chinese, due to its lack of inflection. Specifying the types of extraneous function words into three common types, \emph{Preposition}, \emph{Adverb} and \emph{Particle} could be useful to discuss whether there is difference among these word classes. 

The two other sub-types of \emph{Function word} (\emph{Incorrect} and \emph{Missing}) are not specified in conformity with the initial examination of the data. Not only might adding the extra layer for both sub-error types
not prove practical, but it is also not advised by the MQM guidelines to have the error taxonomy so big that it could challenge annotators' memory limit \cite{burchardt-guidelines-2014}.

\subsection{Results and Discussion}\label{s:result}

\subsubsection{Inter-annotator Agreement}

Inter-annotator agreement (IAA) was calculated with Cohen's Kappa ($\kappa$) \cite{cohen-kappa-1960} on the annotations of the calibration and evaluation sets for the RNN and PATECH's Transformer systems (Table \ref{tab:total_iaa}). 
It is worth noting that the IAA values of the evaluation set improve considerably upon those of the calibration set ($\kappa=0.44$ versus $0.27$). It shows that the discussion of annotation disagreements 
can contribute to improving the level of agreement notably. 

\begin{table}[ht!]
\centering
\small
\begin{tabular}{l|c|m{5em}|c}
\hline
\bf{IAA}  & RNN & Transformer (PATECH) & Both
\\ \hline
Calibration set & 0.31 & 0.22 & 0.27
\\ \hline
Evaluation set & 0.45 & 0.43 & 0.44
\\ \hline
\end{tabular}
\caption{Total and average inter-annotator agreement (Cohen’s κ values) for the MQM calibration set and evaluation set.}
\label{tab:total_iaa}
\end{table}

As shown in Table \ref{tab:total_iaa}, the difference of IAA scores between Transformer and RNN is slight in our evaluation set. The average IAA value (0.44), 
corresponds to moderate agreement, according to Cohen~\shortcite{cohen-kappa-1960}.
When interpreting these results, it should be taken into account that IAA scores are known to be low in human evaluation of MT.  For example, Callison-Burch et al.~\shortcite{callison-burch_meta-_2007} observed fair agreements for fluency and accuracy for eight language pairs, and, though the MQM framework is rigorously defined and supported by clear guidelines, in the experiments by Lommel and Burchardt~\shortcite{lommel-iaa-2014} 
MQM led to relatively low IAA, due to span-level difference, ambiguous categorisation and differences of opinion. 
Klubička et al.~\shortcite{klubicka_quantitative_2018}  reported a moderate agreement on English--Croatian, higher than that by Lommel and Burchardt~\shortcite{lommel-iaa-2014}, probably 
because the agreement was calculated on errors annotated for each sentence, thus not taking the spans of the annotations into account. 
Our own IAA results do not differ greatly with aforementioned research. 

\begin{table}[ht!]
\centering
\begin{tabular}{lc m{5em} c}
\hline 
                           & RNN           & Transformer (PATECH)   & Both          \\
\hline
Accuracy                   & 0.60          & \bf{0.61} & \bf{0.61} \\
\hspace{2mm}Mistranslation & 0.50          & 0.52          & 0.51          \\
\hspace{4mm}Entity         & -0.03         & 0.39          & 0.18          \\
\hspace{4mm}Overly-literal & 0.24          & 0.21          & 0.23          \\
\hspace{2mm}Omission       & 0.52          & \bf{0.67} & 0.60          \\
\hspace{2mm}Addition       & 0.37          & 0.00          & 0.19          \\
\hspace{2mm}Untranslated   & \bf{0.73} & \bf{0.71} & \bf{0.72} \\
Fluency                    & 0.01          & 0.07          & 0.04          \\
\hspace{2mm}Grammar        & 0.36          & 0.24          & 0.30          \\
\hspace{4mm}Function word  & 0.17          & -0.01         & 0.08          \\
\hspace{6mm}Extraneous     & 0.32          & -0.01         & 0.16          \\
\hspace{8mm}Preposition    & \bf{0.65} & -0.01         & 0.32          \\
\hspace{8mm}Adverb         & 0.00          & N/A           & N/A           \\
\hspace{8mm}Particle       & -0.02         & -0.03         & -0.03         \\
\hspace{6mm}Incorrect      & -0.02         & -0.01         & -0.02         \\
\hspace{6mm}Missing        & 0.32          & 0.00          & 0.16          \\
\hspace{4mm}Word order     & 0.45          & 0.29          & 0.37          \\
\hspace{4mm}Classifier     & N/A           & N/A           & N/A           \\
\hspace{2mm}Unintelligible & 0.20          & -0.02         & 0.09          \\
\hspace{2mm}Typography     & 0.22          & 0.28          & 0.25          \\
\hspace{4mm}Punctuation    & 0.21          & 0.29          & 0.25          \\
\hspace{4mm}Unpaired-mark  & N/A           & N/A           & N/A    
\\\hline
\end{tabular}
\caption{Inter-annotator agreement (Cohen’s $\kappa$ values) on the evaluation set for the RNN and PATECH's Transformer systems and their average. Substantial scores (0.61--0.80) are shown in bold. N/A is given to the error categories that were never used, since no data points could be used to calculate the IAA score.}
\label{tab:iaa}
\end{table}

In addition to overall IAA, Cohen's ($\kappa$) was also calculated for each issue type in the evaluation set individually (Table \ref{tab:iaa}). For both systems, the IAA scores for \emph{Accuracy} and its sub-types are considerably higher than those under \emph{Fluency}. It is an expected result taken into account that accuracy errors are more straightforward and less open to interpretation. The $\kappa$ values are relatively consistent between Transformer and RNN, except a striking plunge in agreement scores for Transformer in some categories (
\emph{Function word} and its subtypes, \emph{Word order} and \emph{Unintelligible}) and the opposite, a considerably lower agreement for RNN, for \emph{Entity}.

The source of these disagreements can be traced back to the annotation output.
For example, in the case of \emph{Unintelligible}, the evaluators annotated different sentences with this error category.
As for \emph{Entity}, it is worth mentioning that disagreement arose over this category in the annotation of the calibration set.
It seems that, despite the discussion, the understanding of entity was still not shared by the two annotators.
It is also possible that due to the improved translation quality of Transformer, mistakes such as \emph{Function word} are more subtle and harder to detect.



\subsubsection{Annotated Errors}

Table \ref{tab:total-error} presents the overall number of annotated error tags in the output of each system by each annotator.
One can clearly observe that both annotators have annotated relatively less errors in Transformer's output (PATECH) than in RNN's; the error reduction is of 35\% in the case of annotator 1 and of 27\% in the case of the second annotator.
The Transformer system from KSAI only reduces the number of errors by 12.5\%, compared to the RNN system.


\begin{table}[ht!]
\centering
\small
\begin{tabular}{l|c| m{5em} | m{5em}}
\hline 
System  & RNN & Transformer (PATECH) & Transformer (KSAI)
\\ \hline
Annotator 1 & 168 & 109 & 147
\\ \hline 
Annotator 2  & 193 & 141 & 
\\ \hline
\end{tabular}
\caption{Total amounts of error per annotator and system, as annotated in MQM.}
\label{tab:total-error}
\end{table}

\begin{figure}[ht!]
\centering
\includegraphics[scale=0.45]{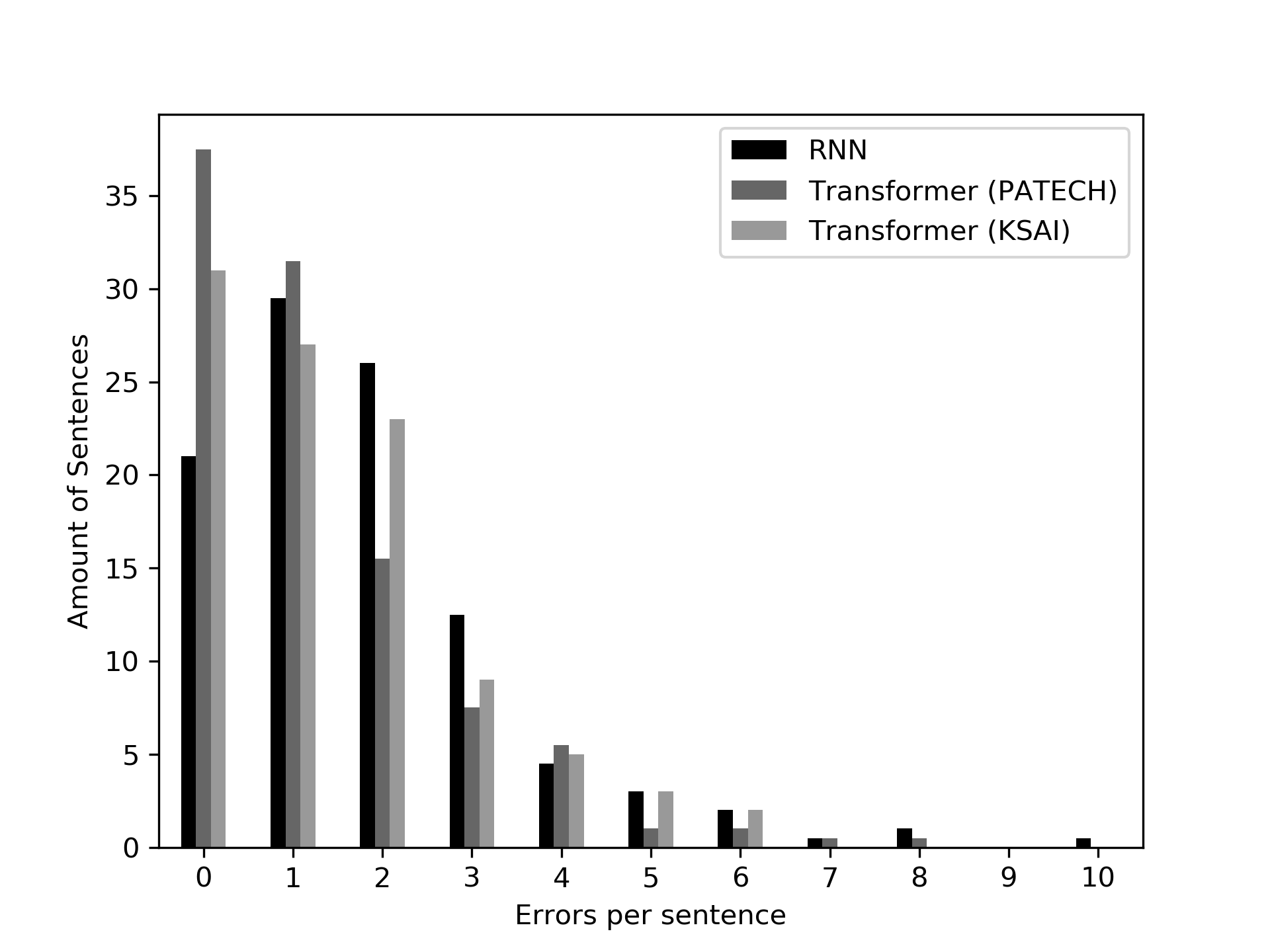}
\caption{Error distribution per system. For RNN and Transformer (PATECH), the average of annotation data from both annotators has been used.}
\label{fig:error per sentence}
\end{figure}

To delve deeper into the error distribution, we plot a histogram to show how many errors appear in each sentence and how many of these sentences are there in the output from each system. The mean of both annotators' annotations for the first two systems are used, amounting to 100 sentences per system. The histogram is shown in Figure~\ref{fig:error per sentence}. It can be observed that more than 35 sentences in the Transformer (PATECH) output are not annotated with any error while only slightly over 20 sentences in the RNN are marked as errorless. The two systems have similar amount of sentences with one mistake, while PATECH's output contains considerably less sentences than RNN with more than one error.

We can also see notable differences between the two Transformer systems in Figure~\ref{fig:error per sentence}. 
Fewer sentences in the KSAI's output are annotated without error, while considerably more sentences are tagged with two errors in this output than in PATECH's system.

 

While comparing the systems in terms of their total number of errors gives us a clear indication of their relative performance, we note that a fairer comparison should take into account their outputs' lengths.
To that end, we make use of the normalisation approach proposed by Klubička et al.~\shortcite{klubicka_quantitative_2018}: 
tokens annotated with errors are counted for each system's output and they are then used to compute each system's error ratio, which equals to the total number of erroneous tokens (Chinese characters) divided by the total number of tokens in the system's output.
This error ratio can serve then as a general score for each system. 
We also apply the same normalisation procedure to each issue type.
Statistical significance for the total amount of errors and each issue type is computed with a pairwise chi-squared (\(\chi^2\)) test \cite{plackett-1983-chi-square}, following its application to normalised MQM errors introduced by Klubička et al.~\shortcite{klubicka_quantitative_2018}.

Table~\ref{tab:word-error-chi-square} shows the error ratios (both overall and for each issue type) for each system, together with an indication of whether there are significant differences between each pair of systems.
In terms of total error ratio, compared to RNN, the error reduction by PATECH amounts to 34\% relative (11.85\% versus 17.93\%) and is significant ($p<0.001$).
No significant difference is observed between the two Transformer-based systems. 
 
For nearly half of the error types,
the decrease in error ratio for Transformer (PATECH), compared to RNN, is statistically significant.
For example, the number of tokens with \emph{Fluency} errors decreased by 45\% (6.45\% verse 3.56\%, $p<0.001$). 
The reduction is particularly notable for its child category \emph{Unintelligible}, for which the number of erroneous tokens decreased by 55\% (2.1\% verse 0.93\%, $p<0.001$). 
This Transformer-based system also managed to generate significantly less extraneous \emph{Function words}, gaining a decrease of 47\% (0.51\% verse 0.27\%, $p<0.05$). 
In addition, Transformer manages to produce significantly less extraneous \emph{Prepositions}, (0.2\% verse 0.04\%, $p<0.05$). 
Though it also produces less \emph{Overly-literal} translations (1.20\% verse 0.86\%) and no extraneous \emph{Adverb} ( 0.06\% verse 0\%), these differences are not significant.

Conversely, this Transformer-based system underperforms on \emph{Punctuation} (0.2\% verse 0.37\%), although the difference is not significant. 
By tracing this back to the annotation, we can observe that Transformer (PATECH) produces several cases of missing, wrong or redundant punctuation marks. For example, in one instance 
an English period (.) was used instead of a Chinese full stop (。).
This Transformer system also had issues with adding guillemets (《》) around newspaper names and putting commas after adverbials,  which are required in Chinese grammar.

Between the two Transformer systems, we can see that except for the category \emph{Entity} and \emph{Untranslated}, the two Transformer systems do not produce statistically significant different amount of errors. It proves that there are few significant discrepancies between these two systems.

Finally, we note that the error category \emph{Unpaired-mark}, has not been used by any of the annotators for any of the three MT systems and the category \emph{Classifier} has only been used to annotate 6 tokens (0.16\%) in the third system's output. While these categories were relevant in MT in the past (see Section~\ref{s:related_work}), our results seem to indicate that they can be considered to have been solved by NMT.

\begin{table}[ht!]
\centering
\begin{tabular}{l|l|m{3em}|m{3em}}
\hline
                           & RNN         & Transf-ormer (PATECH) & Transf-ormer (KSAI) \\
\hline
Accuracy                   & \bf{11.48}       & 8.29**                 & 7.41               \\
\hspace{2mm}Mistranslation & \bf{7.49}        & 4.50**                  & 4.39               \\
\hspace{4mm}Entity         & 0.24        & 0.23                 & \bf{0.59*}               \\
\hspace{4mm}Overly-literal & 1.20         & 0.86                 & 0.51               \\
\hspace{2mm}Omission       & \bf{0.61}        & 0.33**                 & 0.35               \\
\hspace{2mm}Addition       & 0.23        & 0.19                 & 0.22               \\
\hspace{2mm}Untranslated   & 3.16        & \bf{3.27}                 & 2.45*               \\
Fluency                    & \bf{6.45}        & 3.56**                 & 3.02               \\
\hspace{2mm}Grammar        & \bf{3.08}        & 1.83**                 & 2.24               \\
\hspace{4mm}Function word  & \bf{0.51}        & 0.27**                 & 0.40                \\
\hspace{6mm}Extraneous     & \bf{0.35}        & 0.12**                & 0.30                \\
\hspace{8mm}Preposition    & \bf{0.20}         & 0.04**                 & 0.13               \\
\hspace{8mm}Adverb         & 0.06        & 0                    & 0.05               \\
\hspace{8mm}Particle       & 0.07        & 0.08                 & 0.08               \\
\hspace{6mm}Incorrect      & 0.06        & 0.08                 & 0                  \\
\hspace{6mm}Missing        & 0.10         & 0.07                 & 0.11               \\
\hspace{4mm}Word order     & \bf{2.32}        & 1.41**                 & 1.46               \\
\hspace{4mm}Classifier     & 0           & 0                    & 0.16               \\
\hspace{2mm}Unintelligible & \bf{2.10}         & 0.93**                 & 0                  \\
\hspace{2mm}Typography     & 0.20         & 0.37                 & 0.59               \\
\hspace{4mm}Punctuation    & 0.20         & 0.37                 & 0.59               \\
\hspace{4mm}Unpaired-mark  & 0           & 0                    & 0                  \\ \hline
Total error ratio          & \bf{17.93}       & 11.85**                & 10.40   \\\hline
\end{tabular}
\caption{Error ratio (\%) for each error type and overall.
The annotations on RNN and Transformer (PATECH) from both annotators are concatenated.
* indicates \emph{p}-value $<$ 0.05 and ** \emph{p}-value $<$ 0.001, when a system is compared to the system adjacent to its left side. Numbers shown in bold indicate that the system has significantly more erroneous tokens in the pair comparison.}

\label{tab:word-error-chi-square}
\end{table}




\section{Conclusion}\label{s:conclusion}

This paper presented a fine-grained manual evaluation for English$\rightarrow$Chinese on  the two mainstream architectures of NMT: RNN and Transformer. The evaluation was approached in the form of a human error annotation based on a customised MQM error taxonomy.

The error taxonomy was developed from the MQM core taxonomy for MT evaluation. Chinese linguistic features and issues emerged in the calibration set were taken into account by including customised error types, such as \emph{Extraneous function word}, \emph{Classifier} and \emph{Typography}. The error type \emph{Extraneous function word} underpins investigating westernised Chinese phenomena of extraneous function words by specifying it into three word classes: \textit{Preposition}, \textit{Adverb} and \textit{Particle}. 

From our analysis, it is clear that Transformer-based systems generate significantly more accurate, fluent and comprehensible translation with less westernised Chinese expressions. However, Transformer systems do not handle typography as well as RNN. We also note that none of the MT systems did produce any errors related to unpaired-marks and only one system produced errors related to classifiers, which were very unfrequent (0.16\% of the tokens). We can conclude that Transformer systems produce an overall better translation compared to RNN when translating from English to Chinese, which corroborates findings of prior studies on other language pairs. 
A limitation worth mentioning is that our annotation was conducted by only two annotators on a limited amount of data. 

Our taxonomy could be of use for further error analysis on Chinese MT quality. 
Future research could include a larger annotation sample to investigate if punctuation is a a common issue in NMT systems based on Transformer and to verify that NMT is able to produce correct classifiers. Also, as Transformer still shows a major problem in mistranslation, the error taxonomy can be extended with more specific categories to explore this issue in more detail.

The annotations for the three MT systems and the code used for the analysis thereof are publicly available.\footnote{\url{https://github.com/yy-ye/mqm-analysis}}

\section*{Acknowledgements}
We would like to thank Rico Sennrich, for translating the test set used in this paper with a system he had co-developed for a previous edition of the WMT news translation shared task, and Filip Klubi{\v{c}}ka, for providing us with the code to perform the statistical analysis of MQM output.

\bibliography{reference}
\bibliographystyle{eamt20}
\end{CJK*}
\end{document}